%% file: main_cvpr.tex
\documentclass[10pt,twocolumn,letterpaper]{article}

\usepackage{iccv}
\usepackage{times}
\usepackage{epsfig}
\usepackage{graphicx}
\usepackage{amsmath}
\usepackage{amssymb}

\newcommand{\rcite}[1]{\textcolor{red}{\cite{}}}
\usepackage{caption}
\usepackage{stackengine}
\usepackage[caption=false]{subfig}
\usepackage{comment}
\usepackage{booktabs} 
\usepackage{boldline,multirow}
\usepackage[pagebackref=true,breaklinks=true,letterpaper=true,colorlinks,bookmarks=false]{hyperref}

\iccvfinalcopy 

\ificcvfinal\fi
\begin{document}

	\title{Can Synthetic Faces Undo the Damage of Dataset Bias \\To Face Recognition and Facial Landmark Detection?}
	\author{Adam Kortylewski \;\; Bernhard Egger \;\; Andreas Morel-Forster\;\; Andreas Schneider \\ Thomas Gerig \;\; Clemens Blumer \;\;  Corius Reyneke\;\; Thomas Vetter\vspace{10pt}\\ 
		Department of Mathematics and Computer Science  \;\; \\
		University of Basel}

\maketitle

	\begin{abstract}
	It is well known that deep learning approaches to face recognition and facial landmark detection suffer from biases in modern training datasets. 
	In this work, we propose to use synthetic face images to reduce the negative effects of dataset biases on these tasks.  
	Using a 3D morphable face model, we generate large amounts of synthetic face images with full control over facial shape and color, pose, illumination, and background.
    With a series of experiments, we extensively test the effects of priming deep nets by pre-training them with synthetic faces. 
    We observe the following positive effects for face recognition and facial landmark detection tasks: 
	1) Priming with synthetic face images improves the performance consistently across all benchmarks because it reduces the negative effects of biases in the training data. 
	2) Traditional approaches for reducing the damage of dataset bias, such as data augmentation and transfer learning, are less effective than training with synthetic faces. 
	3) Using synthetic data, we can reduce the size of real-world datasets by 75\% for face recognition and by 50\% for facial landmark detection while maintaining performance. 
	Thus, offering a means to focus the data collection process on less but higher quality data. 
	\end{abstract}

\input{introduction}

	\input{related_work}
	\input{face_generator}

	\input{experiments}

	\input{conclusion}

	\bibliographystyle{spmpsci}
	\bibliography{egbib}  

\end{document}

%% file: introduction.tex
    \section{Introduction}
    Facial image analysis tasks such as face recognition and facial landmark detection have gained a lot of attention from the computer vision community. In recent years, advances in deep learning \cite{krizhevsky2012imagenet} and the availability of large-scale datasets led to a great performance increase in face recognition and facial landmark detection \cite{taigman2014deepface,schroff2015facenet,ranjan2017hyperface}. 
    However, the dependence of deep learning approaches on large-scale datasets is a limiting factor because it is  difficult to collect unbiased datasets. In the context of facial image analysis, a key issue is the difficulty of annotating facial properties such as e.g. the head pose, illumination, skin color or facial expression, which in turn makes it difficult to take these properties into account during data collection. Current large-scale datasets were mostly collected from the web and are therefore biased regarding many facial properties. It is well known that such biases have a strong negative influence on the generalization performance of machine learning systems \cite{torralba2011unbiased,ECCV12_Khosla,tommasi2017deeper,kortylewski2017empirically}.
     
   	 Simulated datasets \cite{qiu2016unrealcv,jaderberg2014synthetic,butler2012naturalistic,chen2015deepdriving,gaidon2016virtual,park2015articulated}  
   	 have proven to be reasonably effective for deep learning when real-world training data is scarce, e.g. in the context of autonomous driving \cite{muller2018sim4cv}, indoor scene understanding \cite{jiang2018configurable}, 3D object detection \cite{loing2018virtual} or 3D face reconstruction \cite{kim2017inversefacenet}. It is well known that training from a large-scale synthetic dataset followed by fine-tuning with small set of real-world data is beneficial for deep learning approaches. However, if the dataset used for fine-tuning is also large, deep neural networks tend to forget the information learned from the synthetic data upon learning from the real data. This phenomenon is known as catastrophic forgetting \cite{c1,c2,c3}. Thus, it remains unclear if simulated datasets are useful for facial image analysis tasks for which large-scale real-world datasets are available. 
   	 
    In this work, we study if synthetic face images can be used to improve the generalization performance for two facial image analysis tasks, for which real-world data is abundant: face recognition and facial landmark detection. The largest publicly available datasets for these tasks have severe sampling biases, which limit the generalization performance across different benchmarks \cite{XXX} (also illustrated in Section \ref{sec:exp}). This leads to well-known issues such as a lack of diversity and fairness in face recognition \cite{klare2012face} or the inability to detect facial landmarks for in-the-wild faces, i.e. in rare head poses and illumination conditions \cite{sagonas2013300}. Data synthesis enables the generation of massive amounts of training samples with fine-grained control over well-known biases, such as head pose, the number of facial identities or facial expressions. We propose to combine real world data with synthetic data to overcome the negative effects of dataset biases in face recognition and facial landmark detection. 
	
	For the synthesis of face images we use a 3D Morphable Face Model (3DMM) \cite{blanz1999morphable,gerig2018morphable}. The 3DMM is a statistical model of 3D faces which has been widely applied to facial image analysis in an analysis-by-synthesis setting.  The generation of novel synthetic face images is a basic capability of the 3DMM, which allows us to create datasets with any desired distribution of facial properties that are difficult to annotate in real data (such as illumination, pose, facial identity, facial expression). The advantage of using 3DMMs for data synthesis over related generative face models such as e.g. GANs \cite{berthelot2017began,goodfellow2014generative} is that the 3DMM provides full control over disentangled parameters that change the facial identity in the terms of shape and albedo texture as well as pose, illumination and facial expression. 

	In our experiments, we pre-train deep neural networks with synthetic face images and subsequently fine-tune them with varying amounts of real-world data. Our experimental results reveal the following novel insights about of deep learning approaches for face recognition and facial landmark detection tasks:
    \begin{enumerate}
    	\item \textbf{Enhanced generalization performance due to reduced dataset bias.} The generalization performance of deep neural networks is enhanced consistently across all benchmark datasets (Section \ref{sec:exp-gapclose}), because the negative effects of biases in data are reduced (Section \ref{sec:exp-changingSynthData}).
        \item \textbf{Superiority when compared to data augmentation and transfer learning approaches.} Priming deep neural networks with synthetic faces has a considerably more positive effect on the generalization performance when compared to general techniques for reducing dataset bias, such as data augmentation and transfer learning from classification tasks (Section \ref{sec:exp-face-specific}). 
    	\item \textbf{Enhanced data-efficiency.} The amount of real-world data needed to achieve competitive performance is reduced considerably using synthetic training data (Section \ref{sec:exp-gapclose}). Thus, offering a means for concentrating data collection efforts to less but higher quality data.
    \end{enumerate}
    
	 Curiously, despite the success of 3D Morphable Face Models at facial image analysis, we are not aware of any previous work that uses this effective and easily accessible approach to enhance face recognition or facial landmark detection systems.

%% file: related_work.tex
    \section{Related Work}
    \label{sec:related}  
    
    \textbf{Damage of Dataset Bias.}
    Any data collection is biased
    Since  researchers discovered the presence of a bias in each image data collection 
    classifying caltech-101 e.g. the class fish by looking at the background
    
    \cite{tommasi2017deeper} Transferability of deep features to new domains
    Fine-tuning and catastrophic forgetting \cite{c1,c2,c3}
    \cite{gupta2018robot} biases in robot learning
	\cite{embbias} exposes a offensive associations bias in word embeddings
    \cite{zhang2018examining} diagnose feature representations of the CNNs to discover representation flaws caused by potential dataset bias.
    \cite{mclaughlin2015data} Propose to use data augmentation for reducing the negative effects of dataset bias for person re-identification
    "If you train with biased data, you’ll get biased result"
    In the context of facial image analysis:
    \cite{panda2018contemplating} Identify biases in visual emotion datasets.
    Face recognition, ibm dataset, nist challenges  
    \cite{buolamwini2018gender} gender shades project revealing substantial disparities in error rates based on skin color
       
    \textbf{Deep face recognition.} The performance of face recognition systems significantly increased with the introduction of deep convolutional neural networks \cite{parkhi2015deep,lecun1995convolutional}. In the last years, major performance gains could be achieved by increasing the amount, quality and availability of training data. Taigman et al. \cite{taigman2014deepface} used four million images to train a Siamese network architecture. Further architectural changes have since steadily increased face recognition performances such as DeepID 1-3 \cite{sun2014deep2,sun2014deep1,sun2015deepid3} or the pose-aware model of Masi et al. \cite{masi2016pose}. 
    
    A particularly successful network architecture was proposed by Schroff et al. \cite{schroff2015facenet}, who introduced an innovative triplet-loss to their FaceNet model. 
    For our face recognition experiments we use a publicly available implementation of this FaceNet network \footnote{https://cmusatyalab.github.io/openface/}.
    
    \textbf{Deep facial landmark detection.} 
	The detection of fiducial  points is one of the most important tasks in facial image analysis. Common approaches include regression-based \cite{xiong2013supervised,cao2014face,kazemi2014one} and model-based \cite{cootes1995active,matthews2004active} methods. The former learns local updates given an initial alignment and the latter is based on adapting a holistic face model to the target image. In recent years, facial landmark detection approaches based on deep learning have achieved remarkable levels of performance \cite{sun2013deep}. Trigeorigis et al. \cite{trigeorgis2016mnemonic} combined a CNN with an RNN for cascaded regression and thereby jointly learned a feature representation for each fiducial point as well as the local update procedure. Multi-task learning approaches have gained recent popularity \cite{zhang2014facial,ranjan2017all}. They train a single deep network for the prediction of multiple facial image analysis tasks, thus enforcing the development of a holistic face representation which improves robustness and generalization across all individual tasks.	In this work, we investigate whether deep learning approaches to facial landmark detection benefit from training with synthetic data. For our facial landmark detection experiments we use a publicly available implementation of the multi-task HyperFace network \footnote{https://github.com/takiyu/hyperface}.

    \textbf{Facial image analysis with augmented data.} Massive training datasets have been a critical component for deep learning systems to achieve ever higher performance at facial image analysis tasks. However, the collection, annotation, and publication of millions of face images have proven to be practically unfeasible. Therefore, a major branch of research has evolved around techniques to augment available data in order to increase dataset sizes artificially. Rudimentary approaches to data augmentation are geometric transformations such as mirroring \cite{chatfield2014return,yang2015mirror}, translational shift \cite{levi2015age} and rotation \cite{wolf2011effective,chen2012dictionary}. 
    Patel et al. \cite{patel2011illumination} performed face image relighting in order to make illumination changes between images of the same person more robust. Hu et al. \cite{hu2018frankenstein} composed novel face images by blending parts of different donor face images into a novel image.
    Masi et al. \cite{masi2016we} used domain-specific data augmentation by augmenting the Casia dataset \cite{casia} in 3D. They aligned a 3D face shape to landmarks in each training image and subsequently modified the face with regard to shape deformation, 3D pose, and expression neutralization. This artificial augmentation process does not follow any statistical distribution and thus can lead to unrealistic training data. Furthermore, they did not change the illumination of the images, thus relying on the illumination distribution of the Casia dataset. In this work, we follow a different approach by studying the effect of using fully synthetic face images which were generated with 3DMMs, a statistical illumination model, a camera model and computer graphics.
    
    \textbf{Deep learning with fully synthetic data.} 
    Fully synthetic datasets which are generated with computer graphics have been widely used for the evaluation and training of computer vision tasks such as optical flow \cite{butler2012naturalistic}, autonomous driving systems \cite{chen2015deepdriving}, object detection \cite{gupta2014learning}, pose estimation \cite{shotton2011real,park2015articulated,ionescu2014human3} and text detection \cite{gupta2016synthetic}. Recently, Qiu and Yuille \cite{qiu2017unrealcv} developed UnrealCV, a computer graphics engine for the assessment of computer vision algorithms for scene analysis. Following this approach, Kortylewski et al. \cite{kortylewski2017empirically} recently proposed to use synthetically generated face images for analyzing the generalization performance of different neural network architectures when performing face recognition in the virtual domain. Gaidon et al. presented Virtual KITTI \cite{gaidon2016virtual}, where they used synthetically generated data to pre-train a deep convolutional neural network for the tasks of object detection, tracking and scene segmentation in the context of automated driving systems. They showed that Deep Learning systems behave similarly when trained in the synthetic domain and evaluated in the real domain and vice-versa. In the context of facial analysis, Abbasnejad et al. \cite{abbasnejad2017using} trained a deep convolutional neural network for expression analysis on synthetic data and achieved state-of-the-art results in action unit classification on real data. Kim et al. \cite{kim2017inversefacenet} trained an AlexNet architecture for the regression of 3D morphable model parameters and achieved competitive results when compared to approaches that learn from real-world data. Curiously, despite the widespread use of synthetic data for deep learning, we are not aware of any work that studied how synthetic data can be leveraged to enhance face recognition and facial landmark detection in real-world images.

    In this work, we use a statistical model of 3D face shape, texture, expression and illumination to generate synthetic face images, which we use to train deep neural networks for face recognition and facial landmark detection.

%% file: face_generator.tex
    \begin{figure}
    	\centering
    	\stackunder[2pt]{\includegraphics[width=.15\textwidth]{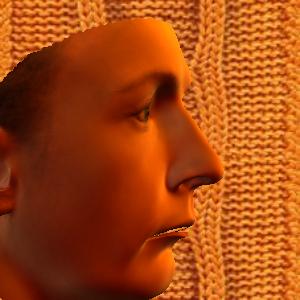}}{}
    	\stackunder[2pt]{\includegraphics[width=.15\textwidth]{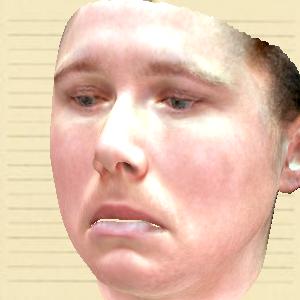}}{}\\
    	\stackunder[5pt]{\includegraphics[width=.15\textwidth]{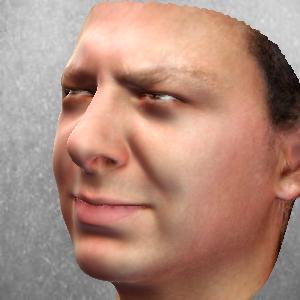}}{}
    	\stackunder[5pt]{\includegraphics[width=.15\textwidth]{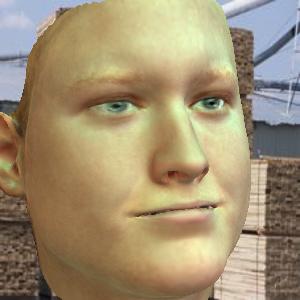}}{}
    	\caption{Example renderings generated with our face image generation process. The facial appearance in the images varies in terms of identity, head pose, expression and illumination. Changes in the background are simulated by overlaying the generated face on random textures.}
    	\label{fig:exampleSynth}
    \end{figure}
	\section{Face Image Generator}
	\label{sec:generator}
	We synthesize face images by sampling from a statistical 3D Morphable Model \cite{blanz1999morphable} of face shape, color and expression. 
	In the following, we describe the most important parameters of the image generation process and their influence on the facial appearance in the image.
	
	\textbf{Facial identity.} In order to simulate different facial identities we use the 
	Basel Face Model 2017 \cite{bfm17} (BFM). The BFM describes a statistical distribution of face shape, color and expression which was learned from 200 neutral high-resolution 3D face scans. The parameters follow a Gaussian distribution. By drawing random samples from this distribution we generate novel 3D face meshes with unique color, shape and facial expression.
	
	\textbf{Illumination.} We assume the Lambertian reflectance model and approximate the environment map with 27 spherical harmonics coefficients (9 for each color channel). 
	 In order to generate natural illuminations, the spherical harmonics illumination parameters are sampled  from the Basel Illumination Prior (BIP) \cite{illuprior}. The BIP describes an empirical distribution of spherical harmonics coefficients estimated from 14'348 real-world face images.
	
	\textbf{Pose and camera}. The simulated faces are viewed with a fixed pinhole camera. The facial orientation with respect to the camera is controlled by six pose parameters (yaw, pitch, roll, and 3D translation). Throughout our experiments, we vary the head pose angles, while normalizing the head position to the center of the image frame.
	
	\textbf{Background}. We simulate changes in the background with a non-parametric background model by sampling randomly from a set of background images of the describable texture database \cite{textures}. The purpose of these random structured background changes is to aid deep learning systems in discovering the irrelevance of background structures for facial image analysis tasks.
	
	The synthesized images (Fig. \ref{fig:exampleSynth}) are fully specified by the aforementioned parameter distributions. Note that our data generation process follows a statistical distribution of face shapes and textures which is in contrast to, for example, the data augmentation in \cite{masi2016we}, where shape deformation between a few fixed 3D shapes is performed. The major benefit of the generator is that by sampling from its parameters we are able to synthesize an arbitrary amount of face images with any desired number of identities, with expressions, in different head poses and natural illumination conditions.

%% file: experiments.tex
    \section{Experiments}
    \label{sec:exp}
    In this section, we empirically analyze the effect of priming deep neural networks with synthetic face images. After introducing our experimental setup (Section \ref{sec:exp-setup}), we analyze the effects of different dataset biases in synthetic and real-world datasets \ref{sec:exp-bias}. We 	study the performance of deep networks when performing face recognition and facial landmark detection trained using a combination of real and synthetic data in Section \ref{sec:exp-gapclose}. We study how biasing the pose distribution and the number of identities in the synthetic data affects the generalization performance of the primed models in Section \ref{sec:exp-changingSynthData}. Finally, we compare our approach to traditional data augmentation methods and to transfer learning from a pre-trained network in Section \ref{sec:exp-face-specific}. 

    \subsection{Experimental Setup}
    \label{sec:exp-setup}
	\textbf{Face recognition.} Our face recognition experiments are based on the publicly available OpenFace framework \cite{openface}. For face detection and alignment we use a publicly available multi-task CNN \footnote{https://github.com/kpzhang93/mtcnn\_face\_detection\\ \_alignment} \cite{mtcnn}. In case the face detection fails, we use the face boxes as defined in the individual datasets \footnote{For LFW and IJB-A these face boxes are provided in the dataset, for Multi-PIE we use the annotations provided in \cite{el2013scalable}.}. We train the FaceNet-NN4 architecture that was originally proposed by Schroff et al. \cite{schroff2015facenet} with the vanilla setting, as provided in the OpenFace framework. The aligned images are scaled to $96\times96$ pixels. The triplet loss is trained with batches of $20$ identities and $15$ sample images per identity for $200$ epochs. 
    
	The real-world training data for face recognition is sampled from the cleaned Casia WebFace dataset \cite{casia}, which comprises 455,594 images of 10,575 different identities. From this dataset, we remove the 27 identities which overlap with the IJB-A dataset. For testing the generalization performance we use: 1) \textbf{CMU-Multi-PIE} \cite{multipie}, which was recorded under controlled illumination and background conditions. We use the neutral identities from session one with the frontal illumination setting. Images from the two overhead cameras are excluded.  2) \textbf{LFW} \cite{lfw} has been the de facto standard face recognition benchmark for many years. Face images in this dataset are subject to a complex illumination, partial occlusion, and background clutter. 3) \textbf{IJB-A} \cite{ijba} was proposed to further push the frontiers of face recognition. The conditions regarding pose, illumination and partial occlusion are more complex than LFW. In addition, subjects are possibly described by multiple gallery images, these image sets are commonly referred to as \textit{templates}. 
	We evaluate face recognition networks at the task of face verification. We measure the distance between two face images as the cosine distance between their 128-dimensional feature embeddings from the last layer of the FaceNet model.
    \begin{equation}
           s(a,b)=\frac{a \cdot b}{\lVert a\rVert_2 \lVert b\rVert_2} \hspace{0.2cm}.
    \end{equation}
    
    For comparing the templates in the IJB-A dataset, we perform softmax averaging of the similarity scores between each image pair as proposed by Masi et al. \cite{masi2016we}. We do not perform any dataset adaptation, thus we test the most challenging face recognition protocol with only \textit{unrestricted, labeled outside data}. For LFW and IJB-A, the pairwise comparisons are provided by their respective protocols. For the Multi-PIE dataset, we follow the LFW protocol. Thus, we perform 10 fold cross validation with 300 pairs of positive and negative samples.
    
    \textbf{Facial landmark detection.} For facial landmarking we use a publicly available implementation  of the HyperFace network \footnote{https://github.com/takiyu/hyperface} \cite{ranjan2017hyperface}. HyperFace is a Multi-Task CNN that learns a common representation in order to perform several facial image analysis tasks jointly: Face detection, pose and gender estimation, and facial landmark detection. For training the HyperFace model, we use the exact training setup as proposed in its open source implementation. Unless otherwise stated, the Neural Networks are randomly initialized with Gaussian noise of zero mean and unit variance.
	
	The real-world data used for training is sampled from the AFLW \cite{koestinger11a} dataset. We randomly select $1K$ images from the AFLW set for testing and use the rest for training. For benchmarking we use three facial landmarking datasets, all of which contain large variations in pose, illumination, facial occlusion and expression: 1) Our test split from the \textbf{AFLW} set. 2) The test set of the \textbf{LFPW} \cite{belhumeur2013localizing} dataset.  3) The \textbf{300-W} \cite{sagonas2013300} dataset which has been introduced as part of a challenge for ''in the wild'' facial landmark detection and is currently one of the most challenging datasets available for facial landmark detection. We combine all 600 images from the indoor and outdoor scenes into our test set.

	In our experiments, we compare the 21 AFLW-landmarks available in AFLW and 300-W. For LFPW we use the 20 landmarks which overlap with the AFLW set. We compute the detection accuracy as the mean euclidean distance between the prediction and the ground truth annotation, normalized according to the face size. 
    \begin{table}
    	\centering
    	\begin{tabular}{l|cV{2.5}cV{2.5}c}
    		\toprule
    		\multicolumn{4}{c}{\textbf{Face Recognition}} \\
    		\hline
    		Datasets & \textbf{Multi-PIE} & \textbf{LFW}  & \textbf{IJB-A} \\
    		\hline  
    		Metric              & Accuracy  & Accuracy  & TAR \\ 
    		\hline  
    		SYN-only            & 88.9     	& 80.1     & 62.5 \\
    		\clineB{1-4}{2.5}
    		Real-100\%          & 91.2     	& 94.1     & 86.8 \\ 
    		\hline  
    		+ Primed      		& \textbf{95.4}     & \textbf{96.0}     & \textbf{92.4} \\
    		\clineB{1-4}{2.5}
    		Real-50\%          	& 86.0     	& 92.7   	& 81.9 \\ 
    		\hline
    		+ Primed  			& \textbf{93.2}    		& \textbf{94.9}     & \textbf{88.7} \\ 
    		\clineB{1-4}{2.5}
    		Real-25\%           	& 83.6     	& 89.1     & 71.3 \\ 
    		\hline 
    		+ Primed  		& \textbf{91.7}    		& \textbf{93.8}     & \textbf{85.6} \\			
    		\clineB{1-4}{2.5}
    		Real-10\%            	& 81.7     	& 85.1     & 66.2 \\ 
    		\hline
    		+ Primed   		& \textbf{90.1}      	& \textbf{92.1}    	& \textbf{83.7} \\			
    		\clineB{1-4}{2.5}
    	\end{tabular}
    	\caption{Face recognition performance on the CMU-Multi-PIE, LFW and IJB-A benchmarks. We compare models trained on synthetic face images (SYN-only) to models trained on different sized subsets of the Casia dataset (Real-$\{10\%,25\%,50\%,100\%\}$). We denote primed models that were fine-tuned on real-world data by ``+ Primed" below the corresponding real-world data only result. We measure performance in terms of recognition accuracy and the true acceptance rate ($TAR$) at false acceptance rate $FAR=0.1$. Priming with synthetic faces improves the face recognition performance considerably.}
    	\label{tab:facerec}
    \end{table}
    \begin{table}
    	\centering
    	\begin{tabular}{c|ccV{2.5}ccV{2.5}cc}
    		\toprule
    		\multicolumn{7}{c}{\textbf{Facial Landmark Detection}} \\
    		\hline
    		Datasets & \multicolumn{2}{cV{2.5}}{\textbf{AFLW}} 	&		\multicolumn{2}{cV{2.5}}{\textbf{LFPW}} &		\multicolumn{2}{c}{\textbf{300-W}}\\				
    		\hline
    		Accuracy      		& 3\%     			& 5\%     			& 3\% & 5\%		  & 3\%      	& 5\%\\
    		\hline
    		SYN only           	& 22.9     			& 67.2      		& 36.9 & 81.6   & 5.1         & 52.5\\
    		\clineB{1-7}{2.5}
    		Real-100\%       	& 47.4     			& 88.8     			& 63.1 & 95.1     & 5.3      	& 78.2\\ 			
    		\hline  
    		+ Primed   			& \textbf{51.8}    & \textbf{89.4}     & \textbf{70.8} & \textbf{95.9}     & \textbf{15.1} & \textbf{88.2}\\
    		\clineB{1-7}{2.5}
    		Real-50\%        	& 43.8     			&  83.5   			& 57.5 & 93.6     & 7.6     	& 75.6\\ 
    		\hline
    		+ Primed    		& \textbf{48.1} 	& \textbf{88.7}     & \textbf{68.9} & \textbf{94.9}     & \textbf{8.8}      	& \textbf{82.7}\\ 
    		\clineB{1-7}{2.5}
    		Real-25\%        	& 35.7     			& 80.8     			& 46.8 & 92.2     & 5.1      	& 70.1\\ 
    		\hline 
    		+ Primed    		& \textbf{42.4}    	& \textbf{85.3}     & \textbf{62.8}  & \textbf{92.8}     & \textbf{6.5}      	& \textbf{77.8}\\			
    		\clineB{1-7}{2.5}
    		Real-10\%        	& 26.5     			& 73.5     			& 29.8 & 85.1     & 5.3    		& 60.7\\ 
    		\hline
    		+ Primed    		& \textbf{38.1}     & \textbf{80.9}    	& \textbf{58.6} & \textbf{91.7}  & \textbf{5.7}      	& \textbf{70.2}\\			
    		\clineB{1-7}{2.5}
    	\end{tabular}
    	\caption{Facial landmark detection performance on the AFLW, LFPW and 300-W benchmarks. We compare models trained on synthetic face images (SYN-only) to models trained on different sized subsets of the AFLW dataset (Real-$\{10\%, 25\%, 50\%, 100\%\}$).  We denote primed models that were fine-tuned on real-world data by ``+ Primed" below the corresponding real-world data only result. We measure detection error at $3\%$ and $5\%$ of the face size (diagonal of the face box). Priming with synthetic faces improves the facial landmark detection performance considerably in all but two experiments.}
    	\label{tab:faceloc}
    \end{table}     
    
    \textbf{Synthetic face image generation.} The synthetic face images used for training are generated by randomly sampling 3DMM parameters as described in Section \ref{sec:generator}. The shape, color and expression parameters are sampled according to the Gaussian distribution defined by the BFM \cite{bfm17}, while the parameters for the illumination are sampled from the empirical BIP \cite{illuprior}. The head pose is sampled according to a uniform pose distribution on the yaw, pitch and roll angles in the respective ranges $r_{yaw}=[ -90^\circ,90^\circ ]$, $r_{pitch}=[ -30^\circ,30^\circ ]$ and $r_{roll}=[ -15^\circ,15^\circ ]$. Background clutter is simulated by randomly sampling a background texture. For face recognition, we generate one million face images with $20K$ different identities and $100$ example images per identity. For landmark detection we use $50K$ training images with $20K$ identities and five sample per identity. The number of synthetic images per task was determined empirically. In Section \ref{sec:exp-changingSynthData} we present an additional experimental analysis of the effect of biasing these dataset characteristics. 	

    \subsection{Dataset Bias in Real and Synthetic Data}
    \label{sec:exp-bias}    
    
	A deep neural networks' ability to generalize to a test set, largely depends on the similarity between the distributions of training and test data (assuming that all hyper parameters and the optimization scheme are the same). Thus, a distributional bias in the training data w.r.t. to a benchmark dataset is directly reflected in the generalization performance. This allows us to measure the presence of biases in the training data for face recognition and facial landmark detection performance (Table \ref{tab:facerec} \& \ref{tab:faceloc}).
	
	When training with the full set of real-world data (Real-100\%) we observe that the distribution of the training data is similar to some benchmark datasets (e.g. LFW \& LFPW), while it is a lot less so for others (e.g. IJB-A \& 300-W). When training with synthetic data only (SYN-only), we observe that for the CMU-Multi-PIE benchmark the performance is similar to that of a deep network trained with real-world data. This suggests that our synthetic face images can well represent the facial appearance in constrained visual environments. However, on the benchmarks of LFW and IJB-A the SYN-only model performs worse when compared to a network trained with the full real-world dataset. Hence, a prominent real-to-virtual performance gap can be observed when testing with ``in the wild" images. 
	
	A similarly prominent real-to-virtual performance gap can be observed for the task of facial landmark detection (Table \ref{tab:faceloc}). The SYN-only model performs worse when compared to the one trained with real-world data (Real-100\%) across all benchmarks.

	These measurements complement a number of previous works which studied the use of synthetic data for training computer vision models. These works show that for image analysis tasks such as e.g. object detection \cite{gupta2014learning} and optical flow \cite{butler2012naturalistic} synthetic data could very well replace real-world data, whereas, for example, for semantic segmentation \cite{mccormac2017scenenet} a real-to-virtual performance gaps exists.
	
	In the context of facial image analysis, it is important to note that biases in the synthetic and real data have very different causes. In real data, some facial properties such as head pose, illumination or facial expression are difficult to annotate and therefore cannot be taken into account when collecting data. In synthetic data, these properties can be modeled very well and thus can be sampled extensively. However, other characteristics of faces are currently not modeled with parametric face models. These include e.g. facial hair, partial occlusion, the mouth interior or detailed skin textures. In the following section, we explore the potential of combining both types of data and their complimentary types of biases.
	
	\begin{table}
		\centering
		\small
		\begin{tabular}{l|cV{2.5}cV{2.5}c}
			\toprule
			\multicolumn{4}{c}{\textbf{Face Recognition}} \\
			\hline
			Datasets & \textbf{Multi-PIE} & \textbf{LFW}  & \textbf{IJB-A} \\
			\hline  
			Metric          & Accuracy  & Accuracy  & TAR \\ 
			\hline  
			Real-100\%      & 91.2     & 94.1     & 86.8 \\ 
			\hline  
			+ Primed  			& 95.4     & 96.0     & 92.4 \\
			\hline     	    		 
			+ Primed-Frontal   	& 91.1     & 93.1     & 82.7 \\ 
			\hline  
			+ Primed-Half    	& 93.5     & 95.6     & 90.2 \\
			\clineB{1-4}{2.5}
		\end{tabular}   	    		
		\caption{Effect of changing the pose distribution and number of facial identities in the synthetic data on the face recognition performance. We measure the recognition accuracy and $TAR$ at $FAR=0.1$ and compare to training with real-world data only (Real-100\%) as well as to priming with the original synthetic data (Primed). The introduction of a strong bias towards frontal face poses of $[ -35^\circ,35^\circ ]$ (Primed-Frontal) reduces the performance. Reducing the amount of facial identities (and thus training data) in the synthetic dataset to $10K$ (Primed-Half) reduces to generalization performance.}
		\label{tab:character-face}
	\end{table}

	\subsection{Priming with Synthetic Data Enhances Performance}
    \label{sec:exp-gapclose}    
	
	We combine synthetic and real-world data by following the common approach of first training the a model with synthetic data followed by a fine-tuning with different subsets (10\%, 25\%, 50\%, 100\%) of the real training data. In this way, the synthetic data primes the model towards the target facial image analysis task, enabling the model to leverage the information in the real-world data more efficiently during the fine-tuning process. The performance of the primed models is denoted as ``+ Primed" in Tables \ref{tab:facerec} \& \ref{tab:faceloc}.	
	
	Note how fine tuning with a small amount of real-world data induces a considerable increase in the performance for both tasks, face recognition and facial landmark detection (Real-10\% + Primed). Thus, confirming the complementary properties of the synthetic and real-world data. This observation is in line with related work on training with synthetic data for general computer vision tasks \cite{} as well as for facial image analysis tasks in particular,  such as XXX \cite{X} or XXX \cite{X}. Importantly, we observe that when more real-world data is used during fine-tuning, the performance keeps increasing. Thus, the models preserve the information learned from the synthetic data and are able to combine it with the additional information from the real data, i.e. we do not observe catastrophic forgetting events. The additional learning effect is most prominent for those benchmark datasets which are most dissimilar from the training data (IJB-A and 300-W). For these we measure a large performance increase for the primed models even when fine-tuning with the full real-world dataset. Thus, the damage of the dataset bias in the large-scale training data can be considerably reduced using the simulated datasets. Note that we do not perform any dataset adaptation in our experiments.

	      \begin{table}
		\centering
		\small
		\begin{tabular}{l|ccV{2.5}ccV{2.5}cc}
			\toprule
			\multicolumn{7}{c}{\textbf{Facial Landmark detection}} \\
			\hline
			Datasets & \multicolumn{2}{cV{2.5}}{\textbf{AFLW}} 	&		\multicolumn{2}{cV{2.5}}{\textbf{LFPW}} &		\multicolumn{2}{c}{\textbf{300-W}}\\				
			\hline
			Accuracy      		& 3\%     & 5\%      & 3\% & 5\%	   & 3\%  & 5\%\\
			\hline
			Real-100\%       	& 47.4    & 88.8     & 63.1 & 95.1     & 5.3  & 78.2\\ 			
			\hline  
			+Primed  			& 51.8    & 89.4     & 70.8 & 95.9     & 15.1 & 88.2\\	
			\hline  
			+Primed-Frontal		& 48.2    & 88.6     & 65.3 & 94.9     & 11.4 & 83.6\\	
			\hline  
			+Primed-Half  		& 51.1    & 89.1     & 70.3 & 95.2     & 14.5 & 87.6\\ 			
			\clineB{1-7}{2.5}
		\end{tabular}
		\caption{Effect of changing the pose distribution and number of facial identities in the synthetic data on the facial landmark detection performance. We compare to training with real-world data only (Real-100\%) as well as to priming with the original synthetic data (Primed). The introduction of a strong bias towards frontal face poses of $[ -35^\circ,35^\circ ]$ (Primed-Frontal) reduces the performance compared to when priming with the unbiased dataset (Primed). Reducing the amount of facial identities (and thus training data) in the synthetic dataset to $10K$ (Primed-Half) reduces to generalization performance.}
		\label{tab:character-loc}
	\end{table}    
     
    In summary, our experimental results of priming deep networks with synthetic faces for face recognition and facial landmark detection (Tables \ref{tab:facerec} \& \ref{tab:faceloc}) demonstrate the following effects:
    \begin{enumerate} 	    
    	\item Priming with synthetic faces followed by fine-tuning with large real-world data leads to a considerable performance increase. 
    	\item The performance increase becomes more prominent when the data distribution in the training and test data are different. Thus, the damage of dataset bias in the training data is considerably reduced.
     	\item After synthetic priming, only $25\%-50\%$ of the real-world data is needed to achieve the same performance as is obtained when training an unprimed model with the full real-world dataset.
    \end{enumerate}
    
	\subsection{Biasing the Synthetic Data Distribution Decreases Performance}
	\label{sec:exp-changingSynthData}
	In this section, we evalute how the additional variability of facial properties in the synthetic data affects the generalization performance of the primed models when performing face recognition and facial landmark detection tasks. In particular, we introduce well-known real-world biases such as a bias to frontal poses and a reduced variability in the number of facial identities in the synthetic dataset. Note that we change one dataset characteristic at a time while keeping all other parameters fixed to the original values of our previous experiments. The experimental results are summarized in Tables \ref{tab:character-face} \& \ref{tab:character-loc}.     
	
	    \begin{table}
	    	\centering
	    	\small       	
	    	\begin{tabular}{c|ccV{2.5}ccV{2.5}cc}
	    		\toprule
	    		\multicolumn{7}{c}{\textbf{Face Recognition}} \\
	    		\hline
	    		Datasets & \multicolumn{2}{cV{2.5}}{\textbf{Multi-PIE}} 	&		\multicolumn{2}{cV{2.5}}{\textbf{LFW}} &		\multicolumn{2}{c}{\textbf{IJB-A}}\\				
	    		\hline
	    		\# real data 		& 50\%     			& 100\%     		& 50\% & 100\%		  	& 50\% & 100\%\\
	    		\hline
	    		Random           	& 86.0     			& 91.2      		& 92.7 & 94.1   		& 81.9  & 86.8\\
	    		\clineB{1-7}{2.5}
	    		Augmented     	    & 87.7     			& 91.3     			& 91.3 & 94.6     	 	& 82.8 	& 86.7\\ 			
	    		\clineB{1-7}{2.5}
	    		Primed        			& \textbf{93.0} 				& \textbf{93.3} 				& \textbf{94.8} & \textbf{95.8}     		& \textbf{88.2} 	& \textbf{90.6}\\ 
	    		\clineB{1-7}{2.5}
	    	\end{tabular}
	    	\caption{Comparison of the face recognition performance when initializing the weights of our model randomly to training with augmented data, and to priming with synthetic face images. We measure the performance when using $50\%$ and $100\%$ of the cleaned Casia dataset for training. We test using the CMU Multi-PIE, LFW and IJB-A benchmarks. While the effect of data augmentation is marginal, priming with synthetic face images leads to a consistently enhanced generalization performance.}
	    	\label{tab:facerec-imp}
	    \end{table}  
	    
	\textbf{Bias to frontal pose.} We bias the yaw pose in the synthetic data to a range of $r_{yaw}=[ -35^\circ,35^\circ ]$, while keeping the amount of data the same as in the original SYN dataset. In our experiments, we prime the models with the biased dataset and subsequently fine-tune with the full real world datasets (``primed-frontal" in Tables \ref{tab:character-face} \& \ref{tab:character-loc} ). From the results, we can observe a performance decrease at both facial image analysis tasks compared to priming with the original unbiased dataset (``primed"). This negative effect is particularly prominent for the most challenging datasets, IJB-A and 300-W. This decrease in the generalization performance demonstrates that deep neural networks do benefit from the large pose variation in the synthetic data, suggesting that conceptual knowledge about 3D head pose variation is learned and transferred to the real-world data. 
  
    \begin{table}
    	\centering
    	\small	    
    	\begin{tabular}{c|ccV{2.5}ccV{2.5}cc}
    		\toprule
    		\multicolumn{7}{c}{\textbf{Facial Landmark Detection}} \\
    		\hline
    		Datasets & \multicolumn{2}{cV{2.5}}{\textbf{AFLW}} 	&		\multicolumn{2}{cV{2.5}}{\textbf{LFPW}} &		\multicolumn{2}{c}{\textbf{300-W}}\\				
    		\hline
    		\# real data 		& 50\%     			& 100\%     		& 50\% & 100\%		  	& 50\% & 100\%\\
    		\hline
    		Random           	& 43.8     			& 47.4      		& 57.5 & 63.1   		& 75.6  & 78.2\\
    		\clineB{1-7}{2.5}
    		Augmented     	    & 44.9     			& 49.7     			& 61.2 & 68.8     		& 74.3 	& 84.3\\ 			
    		\hline  
    		ImageNet  			& 46.3    			& 47.6    			& 65.2 & 65.3     		& 79.8  & 85.1\\ 			
    		\clineB{1-7}{2.5}
    		Primed        			& \textbf{47.8} 	& \textbf{50.9} 	& \textbf{68.5} & \textbf{70.5}     		& \textbf{82.2} 	& \textbf{87.8}\\ 
    		\clineB{1-7}{2.5}
    	\end{tabular}
    	\caption{Comparison of the facial landmark detection performance when initializing the weights of our model randomly to training with augmented data, transfer learning from the ImageNet classification dataset, and to priming with synthetic face images. We measure the performance when using $50\%$ and $100\%$ of the AFLW training data. For the AFLW and LFPW datasets we show the detection accuracy at $3\%$ error and for 300-W at 5\% error. Priming with synthetic data leads to a consistent superior performance.}
    	\label{tab:faceloc-imp}
    \end{table}       
	
	\textbf{Decreasing the number of facial identities.} We reduce the number of identities in the synthetic data to $10K$, while retaining the amount of training images for each facial image analysis task. After priming with the new dataset and subsequently fine-tuning on the full real-world datasets, we observe a decrease in the generalization performance when performing either facial image analysis task (``primed-half" in Tables \ref{tab:character-face} \& \ref{tab:character-loc} ). The performance decrease is particularly prominent for face recognition, whereas for facial landmark detection it is marginal, suggesting that more complex facial image analysis tasks benefit more from additional synthetic data. These results confirm our observations that the additional variability in the synthetic data causes the increased generalization performance.

	In summary, the experiments in this section demonstrate that the additional variability in the synthetic data, in terms of pose and facial identities, is an important factor for the overall increase in the generalization performance of real-world data. Thus, demonstrating that the deep models can retain the complementary information of the simulated when fine-tuned on large real-world datasets.

    \subsection{Comparison to Data Augmentation and Transfer Learning}	
    \label{sec:exp-face-specific}
   
   In this Section, we compare the effectiveness of our approach to related techniques for enhancing the generalization ability of deep neural networks when performing facial image analysis. One popular approach to account for biases in the training data is data augmentation. Thereby, the original training samples are manipulated with general image transformations such as color changes, 2D rotations and mirroring \cite{chatfield2014return,yang2015mirror,wolf2011effective,chen2012dictionary}. Another widely applied tool for performance increase is transfer learning from image classification, where a model is pre-trained on a general object classification task and subsequently fine-tuned on the desired facial image analysis task. In this way, the model can re-use the knowledge acquired for discriminating objects and transfer it to facial image analysis.

   In our experimental setup, we augment training data by mirroring across the vertical axis and by rotating a face image twice around the center of the face randomly in a range of $[-30^\circ,30^\circ]$. In this way, we extend the head pose distribution in the training data. For the pre-training experiments, we use an AlexNet \cite{krizhevsky2012imagenet,jia2014caffe} that was trained for image classification on the ImageNet dataset~\cite{russakovsky2015imagenet}. We subsequently fine-tune this model for facial landmark detection. We could not test this setup for face recognition, as there are no pre-trained FaceNet architectures available. As a general performance baseline, we use a model with randomly initialized weights.
   
   In Tables \ref{tab:facerec-imp} \& \ref{tab:faceloc-imp}, we summarize the performance results for all tested approaches when using $50\%$ of the real-world training data and when using all of the available data for fine-tuning. We observe that data augmentation with 2D image transformations has only a marginal effect on face recognition. Apparently, the large-scale face recognition dataset already offers enough data to generalize well across the head poses that can be simulated by data augmentation. For facial landmark detection we observe a positive effect, due to the smaller scale of the training dataset. In addition, pre-training seems to have a more positive effect than data augmentation, when only $50\%$ of the training data is used.
   
	Our approach of priming the models with synthetic face images is superior to data augmentation and transfer learning from a classification task. In particular, the performance of face recognition and facial landmark detection is enhanced with any amount of available real-world training data. The major advantage of our approach is that the synthetic data offers \textit{face-specific} knowledge that goes beyond the information offered by general techniques such as data augmentation and transfer learning. Also note that all three approaches are to some extent complementary, and can thus be combined.

%% file: conclusion.tex
    \section{Discussion}
    \label{sec:conclusion}
	In this work, we demonstrated that priming deep neural networks with synthetic face images considerably reduces the negative effects of dataset bias on face recognition and facial landmark detection. In particular, our study provides the following insights:
	
	\textbf{Improved generalization performance due to additional variability in the synthetic data.} Priming with synthetic data followed by fine-tuning with real-world data enhances the generalization performance consistently across all benchmark datasets compared to training with real-world data only. Our experiments in Section \ref{sec:exp-changingSynthData} show that the additional variability in the synthetic data, in terms of head pose and facial identity, is a crucial factor for the measured increase in generalization performance. Thus, we provide evidence that the negative effects of biases in the real-world data are alleviated when priming with synthetic data that is unbiased in these variables.
	
	\textbf{Fine-tuning with large real-world datasets does not induce catastrophic forgetting.} We have not observed catastrophic forgetting effects after fine-tuning. Thus, deep networks can retain the information learned from the simulated data after being fine-tuned on large real-world datasets, while still being able to learning the complementary information available in the real data.
	
	\textbf{The size of real-world datasets can be considerably reduced using synthetic data.} 
	Only relatively few real-world data are needed to learn the image variation which currently cannot be simulated well. Using our priming approach the number of real-world data needed to achieve competitive performance was reduced by 75\% for face recognition, and by 50\% for facial landmark detection.
	
	\textbf{Superiority to data augmentation and transfer learning from classification tasks.} 
	Our experiments show that the proposed method of priming with synthetic face images has a considerably more positive effect on the generalization performance of deep neural networks than data augmentation and transfer learning from a classification task. This observation is intuitive because our synthetic data provides additional information that is specific to the target facial image analysis tasks. This kind of knowledge cannot be learned from a general object classification task nor when the training data is augmented with basic image transformations.
	
	In summary, our experimental results suggest that researchers working in the field of face recognition and facial landmark detection should consider priming their models with synthetic faces for enhanced performance, in particular when the facial properties in the training and test data are expected to be distributed differently. Our data generator and any other software used in this work are publicly available.